\PassOptionsToPackage{table}{xcolor}
\documentclass[10pt,twocolumn,letterpaper]{article}

\usepackage{cvpr}              

\newcommand{\first}[1]{{\textbf{#1}}}
\definecolor{colorfirst}{RGB}{255, 204, 204}
\definecolor{colorsecond}{RGB}{255, 230, 204}
\definecolor{colorthird}{RGB}{255, 251, 214}
\newcommand{\cellfirst}{{\cellcolor{colorfirst}}}
\newcommand{\cellsecond}{{\cellcolor{colorsecond}}}
\newcommand{\cellthird}{{\cellcolor{colorthird}}}

\usepackage{changes}

\newcommand{\figref}[1]{Fig.~\ref{#1}}
\newcommand{\tabref}[1]{Tab.~\ref{#1}}
\newcommand{\secref}[1]{Sec.~\ref{#1}}

\definecolor{cvprblue}{rgb}{0.21,0.49,0.74}
\usepackage[pagebackref,breaklinks,colorlinks,allcolors=cvprblue]{hyperref}
\usepackage{graphicx}
\def\paperID{3222} 
\def\confName{CVPR}
\def\confYear{2025}

\title{AnimateAnything: Consistent and Controllable Animation for Video Generation}

\author {
    Guojun Lei\textsuperscript{\rm 1}\footnotemark[1],    
    Chi Wang\textsuperscript{\rm 1}\footnotemark[1],
    Hong Li\textsuperscript{\rm 3,5}\footnotemark[1],
    Rong Zhang\textsuperscript{\rm 4},
    Yikai Wang\textsuperscript{\rm 2},
    Weiwei Xu\textsuperscript{\rm 1}\footnotemark[2] \\
    \textsuperscript{\rm 1} State~Key~Lab~of~CAD\&CG,~Zhejiang~University \quad     \textsuperscript{\rm 2} Tsinghua University \\ \quad \textsuperscript{\rm 3} Beihang University \quad \textsuperscript{\rm 4} Zhejiang Gongshang University \quad  \textsuperscript{\rm 5} ShengShu \\
    {{{\tt\small \href{mailto:guojunlei@zju.edu.cn}{guojunlei@zju.edu.cn}, \href{mailto:wangchi1995@zju.edu.cn}{wangchi1995@zju.edu.cn}, \href{mailto:link0502@buaa.edu.cn}{link0502@buaa.edu.cn}}}} \\[0em]
    {{\tt\small \href{mailto:zhangrong@zjgsu.edu.cn}{zhangrong@zjgsu.edu.cn}, \href{mailto:yikaiw@outlook.com}{yikaiw@outlook.com}, \href{mailto:xww@cad.zju.edu.cn}{{x}ww@cad.zju.edu.cn}}
    }
}

\begin{document}

\twocolumn[
{
\renewcommand\twocolumn[1][]{#1}
\maketitle
\begin{center}
     \centering
     \vspace{-0.25cm}
        \captionsetup{type=figure}
        \includegraphics[width=1.0\linewidth]{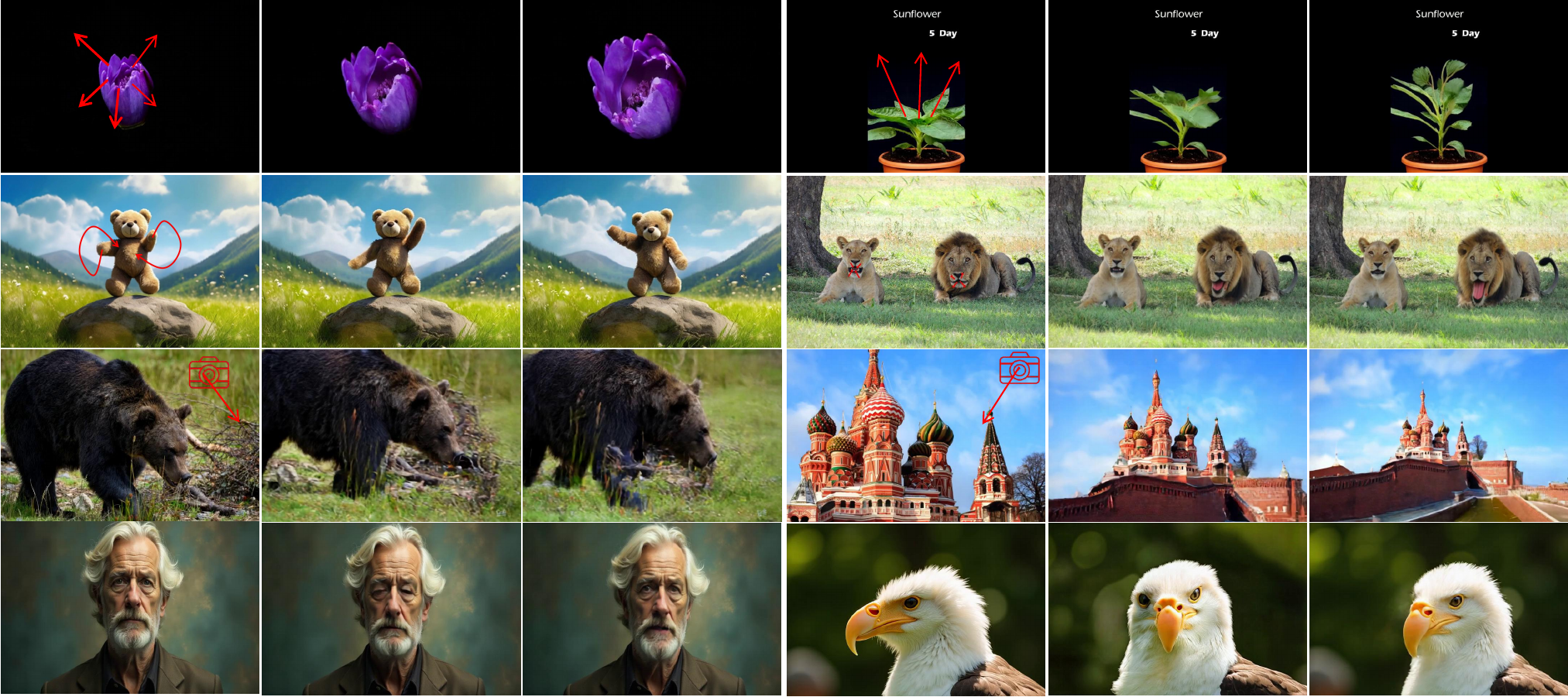}
        \vspace{-0.5cm}
        \captionof{figure}{
        Animate anything. Consistent and controllable animation for different kinds of control signals. Given a reference image and corresponding user prompts, our approach can animate arbitrary characters, generating clear stable videos while maintaining consistency with the appearance details of the reference object.
        }
        \vspace{2em}
        \label{fig:intro}
\end{center}%
}
]

{
  \renewcommand{\thefootnote}{\fnsymbol{footnote}}
  \footnotetext[1]{Joint first authors.}
  \footnotetext[2]{Corresponding author.}
}


\begin{abstract}
We present a unified controllable video generation approach AnimateAnything that facilitates precise and consistent video manipulation across various conditions, including camera trajectories, text prompts, and user motion annotations. Specifically, we carefully design a multi-scale control feature fusion network to construct a common motion representation for different conditions. It explicitly converts all control information into frame-by-frame optical flows. Then we incorporate the optical flows as motion priors to guide the final video generation. In addition, to reduce the flickering issues caused by large-scale motion, we propose a frequency-based stabilization module. It can enhance temporal coherence by ensuring the video's frequency domain consistency. Experiments demonstrate that our method outperforms the state-of-the-art approaches.
For more details and videos, please refer to the anonymous webpage: \url{https://yu-shaonian.github.io/Animate_Anything/}.


\end{abstract}

\section{Introduction}
\label{sec:intro}
The emergence of Sora~\cite{videoworldsimulators2024} has led to a breakthrough in large-scale video generation. Recently, controllable video generation~\cite{wang2024motionctrl,he2024cameractrl,yang2024cogvideox,hu2023animateanyone},~\ie controlling camera trajectories and object movements, has gained significant attention. It has expanded applications of video generation, making them directly applicable to film production and virtual reality. However, due to the high complexity of large-scale camera and object movements, achieving precise control over video generation in such cases remains challenging.



MotionCtrl~\cite{wang2024motionctrl} and CameraCtrl~\cite{he2024cameractrl} support camera trajectory manipulation for dynamic video generation, but they rely solely on text input. Since text descriptions provide only the overall characteristics of a video and cannot convey specific details precisely, it is insufficient to manipulate the video generation process only using text prompts.
In contrast, image guidance, such as user-annotated trajectories or reference videos, can present more detailed visual cues.
Motion-I2V~\cite{Shi_Huang_Wang_2024} allows for image-based guidance but only enables slight object movements through user drag annotations, such as adjusting eye direction or indicating leg motion. It is not capable of manipulating the camera trajectory of a video. 
MOFA-Video~\cite{niu2024mofa} achieves control over detailed, pixel-level movements but is also limited to small-scale camera movement. To ensure global consistency in dynamic videos, MOFA-Video requires users to specify the movement direction for each local region of the input image that may move, making the process overly complex for user interaction.

This paper focus on image-to-video (I2V) generation that simultaneously processes dynamic control signals, such as arrow-based motion annotations, camera movements, and reference videos.
However, the integration of these signals is challenging due to their different modalities, making the direct combination with a single video generation model difficult.
Current approaches~\cite{Animatediff} typically attempt to train each control signal individually (for instance, through Lora \cite{J._Shen_Wallis_Allen-Zhu_Li_Wang_Chen_2021}), and then collaboratively apply these signals to enhance video generation results. Nevertheless, these methods often necessitate careful parameter tuning or denoising strategies, making it difficult to maintain video stability, which can lead to flickering effects or incoherent pixel motion caused by different control signals.
Some methods~\cite{niu2024mofa, Shi_Huang_Wang_2024} aim to facilitate the controllable generation of local motion in videos by introducing optical flow fields; however, they are ineffective in addressing camera motion signals, as camera movement introduces global motion information that is independent of the subject's motion.
Based on the insights presented above, we speculate that if the local motion of the subject and the global motion of the camera can be unified into a representation of frame-by-frame pixel movements, namely optical flow, it would support the guidance of the video generation model's behavior, thereby possessing the potential to achieve synchronous control of various signals.

\begin{figure}[t]
    \includegraphics[width=\linewidth]{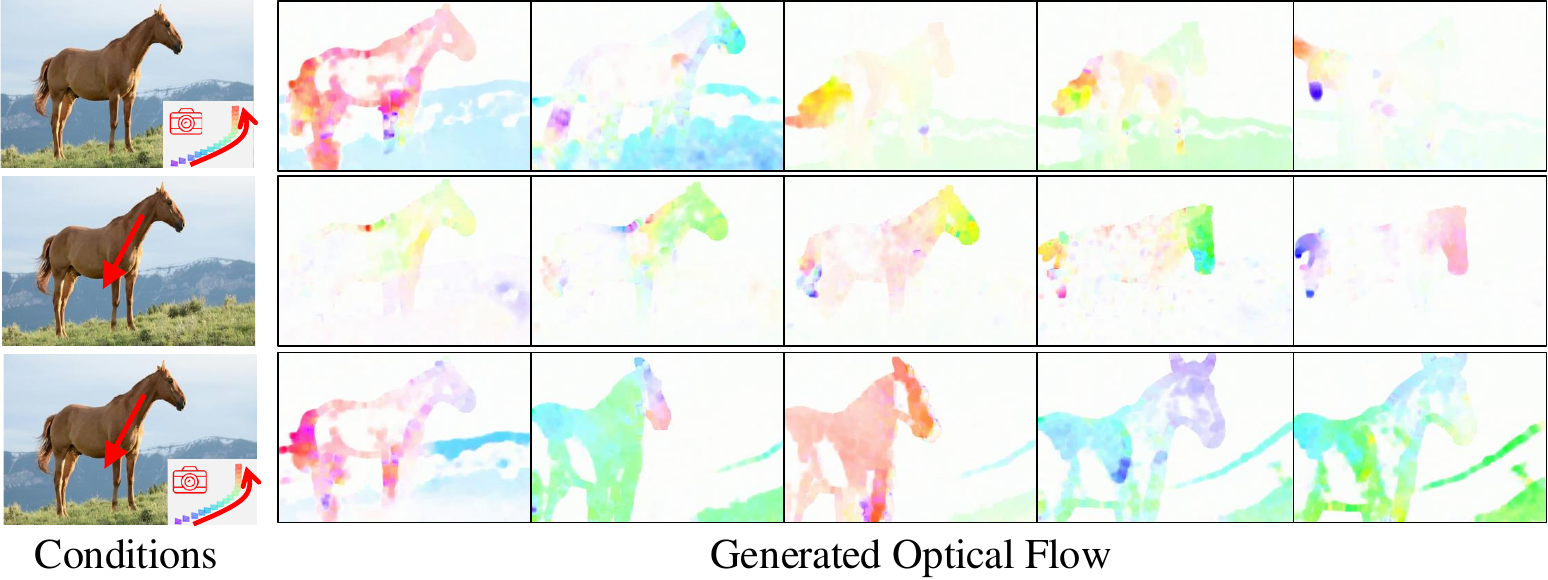}
    \vspace{-2em}
    \caption{The generated optical flow by our method with different condition signals. Given a specific image, from top to bottom are optical flows generated with camera trajectory, arrow-based motion annotation, and both conditions, respectively.}
    \label{fig:conditioanlof}
\end{figure}

The key challenge is to handle incoherent pixel motion caused by different control signals. For instance, as \figref{fig:conditioanlof} shows, 
the optical flow generated from different control signals varies greatly. Camera motion involves global movement, affecting both foreground and background pixels~(the first row of~\figref{fig:conditioanlof}), while a single motion annotation mainly influences localized foreground pixel motion~(the second row of~\figref{fig:conditioanlof}). 
Simultaneously introducing these two conditions directly may result in control signal conflicts, making the model confused about the following movements~(the third row of~\figref{fig:conditioanlof}). It is difficult to directly integrate these conditions through existing methods, since it demands considering both the individual impact of each condition and their complex interactions, such as projection transformations and occlusion completion. Therefore, we strategically design various condition-injection modules based on the representation and correlations of different control signals to enable unified optical flow generation.

Therefore, we propose an innovative two-stage video generation method to achieve multi-condition joint control.
In the first stage, we convert various motion control signals into a unified optical flow, which is then used to guide the final video generation in the second stage.
To further enhance video stability, we convert features from the time domain to the frequency domain and introduce a spectral attention mechanism to improve the overall quality of the generated videos.

The main contributions are summarized as follows:
\begin{itemize}

\item We introduce a two-stage pipeline to achieve stable and flexible video generation with different kinds of control signals. In the first stage, all control signals will be unified into frame-by-frame optical flow, which is then fed into the second stage to synchronize with text controls for high-quality video generation.

\item We utilize an adaptive feature refinement in the frequency domain. This operation effectively suppresses instability and flickering in the generated video by modifying the temporal frequency features within the video.
\item We perform extensive experiments to demonstrate the superiority of our method over state-of-the-art methods both quantitatively and qualitatively.
\end{itemize}

\section{Related Work}
\label{sec:related}

\noindent\textbf{Controllable Video Generation.}
Text-to-video~(T2V) generation has received significant attention in recent years, especially after the emergence of Sora~\cite{J._Shen_Wallis_Allen-Zhu_Li_Wang_Chen_2021}. Typically, in this area, the text-based control information is injected via cross-attention mechanisms~\cite{wang2024motionctrl, he2024cameractrl}. While MagicTime~\cite{yuan2024magictime} introduces a novel GPT~\cite{achiam2023gpt}-based ``Magic Text-Encoder'' to enhance text comprehension ability.
However, text-driven approaches often fail to convey video details precisely. As a result, methods driven by text and image simultaneously have become popular, significantly addressing these limitations. To achieve more effective video generation, pioneers~\cite{wang2024videocomposer, esser2023structure, chen2023seine, zhang2023controlvideo, jain2024peekaboo} have explored generating videos under the guidance of some easily obtainable signals such as edges, depth, optical flow, or bounding boxes. 
Taking reference video as motion guidance, MotionClone~\cite{ling2024motionclone} enables motion cloning by treating temporal-attention weights as motion representation.
Recently, with the immense potential of the film industry and virtual reality applications, precise control over camera and object motion trajectories has gained increasing interest.

\noindent\textbf{Camera Trajectory Driven Video Generation.}
To facilitate camera trajectory control, AnimateDiff~\citep{Animatediff}
trains additional motion LoRA~\cite{hu2021lora} modules for each specific camera path.
However, this method lacks precise control of camera trajectory and cannot generate videos for unseen camera trajectories.
A straightforward solution to these issues is to treat camera parameters as additional conditions for video generation. MotionCtrl~\cite{wang2024motionctrl} employs 12 pose matrix parameters as frame-level conditions to explicitly introduce camera trajectory. Nevertheless, this method still shows limitations in capturing the necessary geometric information for precise camera control. CameraCtrl~\cite{he2024cameractrl} enhances camera information integration by using pl\"ucker embeddings to represent camera trajectories.
To integrate the camera embedding more effectively, VD3D~\cite{vd3d} and CamCo~\cite{xu2024camco} introduce a ControlNet~\cite{Zhang_Agrawala}-like conditioning mechanism and an epipolar attention module, respectively.

\noindent\textbf{Object Motion Trajectory Driven Video Generation.}
CameraCtrl~\cite{he2024cameractrl} and MotionCtrl~\cite{wang2024motionctrl} support basic motion control through text description, which is rough and imprecise. Compared to T2V’s object motion control based on motion descriptions, the Image-to-video~(I2V) method generates videos from object motion trajectories. It allows for a more precise and user-friendly description of object positions, movement directions, and motion amplitude within the scene. 
\citeauthor{yin2023dragnuwa}, \citeauthor{hao2018controllable} and \citeauthor{Shi_Huang_Wang_2024} introduce explicit optical flow as an intermediate representation to guide video generation. Similarly, MOFA-Video takes sparse motion hints as input and generates dense optical flows to warp multi-scale features to guide video generation. 

Motion-I2V~\cite{Shi_Huang_Wang_2024} is the most relevant method to ours, as both use a two-stage framework and select optical flow as an intermediate motion representation. In the first stage of optical flow generation, Motion-I2V derives optical flow from text and reference motion field images. In contrast, we treat optical flow as a unified visual motion representation that incorporates reference images, drag actions, and viewpoint changes. This allows for a more comprehensive and balanced handling of diverse visual conditions. In the second stage, while Motion-I2V directly integrates optical flow with text and images, we employ the Expert AdaLN~\cite{yang2024cogvideox} to promote feature space alignment adaptively, enabling a deep fusion of multimodal features. This enhancement improves the capability and generalization of this stage, allowing it to perform effectively even when the image and optical flow are not fully aligned.


\section{Methods}

\begin{figure*}[t]
    \centering
    \includegraphics[width=1\linewidth]{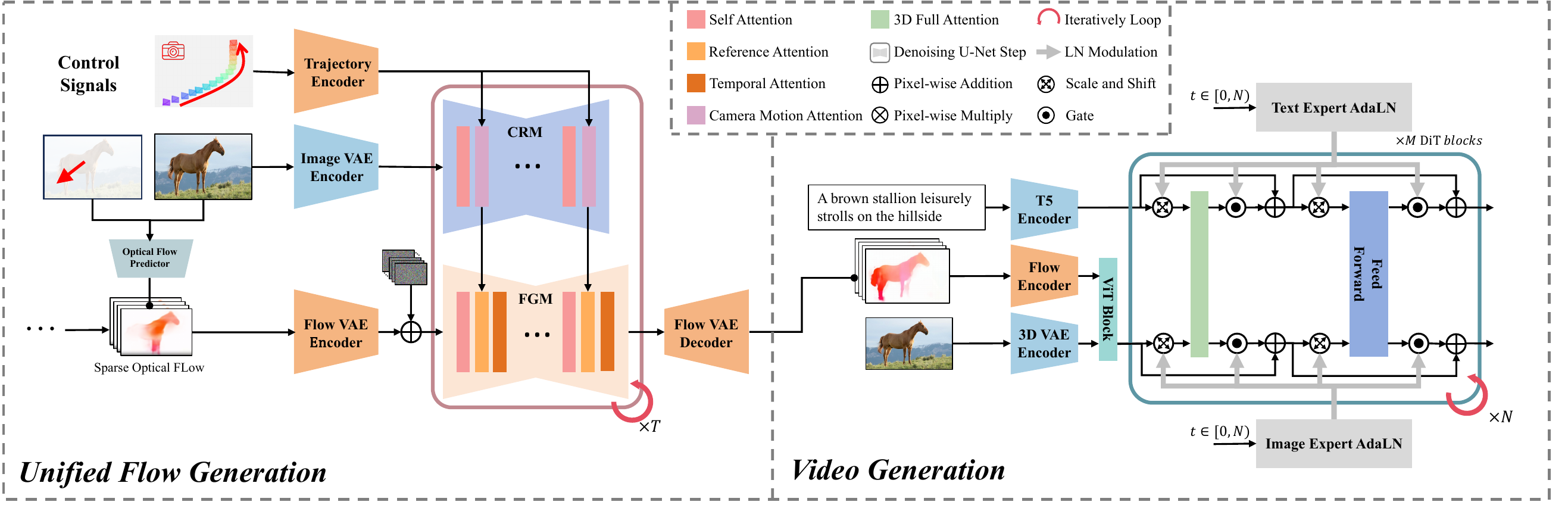}
    \vspace{-1.0em}
    \caption{\textbf{AnimateAnything Pipeline.} 
    The pipeline consists of two stages: 
    1) Unified Flow Generation, which creates a unified optical flow representation by leveraging visual control signals through two synchronized latent diffusion models, namely the Flow Generation Model~(FGM) and the Camera Reference Model~(CRM). The FGM accepts sparse or coarse optical flow derived from visual signals other than camera trajectory. The CRM inputs the encoded reference image and camera trajectory embedding to generate multi-level reference features. These features are fed into a reference attention layer to progressively guide the FGM's denoising process in each time step, producing a unified dense optical flow.
    2) Video Generation, which compresses the generated unified flow with a 3D VAE encoder and integrates it with video latents from the image encoder using a single ViT block. The final output is then combined with text embeddings to generate the final video using the DiT blocks.
    }
    \label{fig:pipeline}
\end{figure*}

In this section, we present AnimateAnything,  a unified controllable video generation approach for precise and consistent video customization across various conditions. As the pipeline illustrated in \figref{fig:pipeline}, 
we convert all visual control signals into a unified optical flow representation and then utilize it to guide the final video generation. 
In the following subsections, we will provide a detailed explanation of preliminary knowledge, each module of the pipeline, and the corresponding training strategies.
Firstly, we provide a brief overview of Video Diffusion Models in \secref{3.1}. 
Afterward, we present the architecture of converting all control signals into unified flows in \secref{3.2}. 
Then, we introduce how the flows guide the final video generation in \secref{3.3}. 
Finally, we give detailed descriptions of the frequency stabilization module~\ref{3.4} and training strategy~\ref{3.5}.

\subsection{Video Diffusion Models} 
\label{3.1}
The video diffusion model builds on the concept of image diffusion probabilistic models, extending into the temporal dimension. It captures the dynamic relationships between frames in a video sequence, allowing for the generation of continuous and high-quality video content. By learning to reverse the added noise, it ensures temporal consistency and coherence in the generated videos.
Let $x_0 \in R^{f\times h\times w \times c}$ represent a video latent variable, where $f$ is the total number of frames, each of size $h \times w$ with $c$ channels. The forward diffusion process is modeled as a chain that incrementally adds Gaussian noise to the original video, defined as follows:
\begin{equation}
x_t=\sqrt{\bar{\alpha_t}} x_{t-1}+\sqrt{\left(1-\bar{\alpha_t}\right)} \epsilon, \epsilon  
 \sim N\left(0,1\right),
\end{equation}
where $t \in \{1, \ldots, T\}$ denotes the timestep, $\bar{\alpha_t}$ regulates the intensity of noise added at each $t$ , and $\epsilon$ is drawn from standard Gaussian noise. 
In the reverse process, a denoising model is learned to estimate $p\left(x_{t-1}|x_t\right)$, typically parameterized by a neural network $\theta$. The optimization objective is to minimize the following loss function:
A denoising model, parametrized by neural network $\theta$, estimates $p\left(x_{t-1}|x_t\right)$ in the reverse process,  minimizing the given loss function:
\begin{equation}
    \mathcal{L}(\theta)=\mathbb{E}_{x_{0}, \epsilon, \mathcal{C}, t}\left[\| \epsilon -\dot{\epsilon}_{\theta}\left(x_{t}, \mathcal{C}, t\right)\|_2^2\right],
\end{equation}
where $\mathcal{C}$ denotes the guidance conditions, like text. To train a video generation diffusion model using images, the image encoding is typically concatenated with the $x_t$, enabling the model to efficiently use its semantic features.

\subsection{Stage 1: Unified Flow Generation}
\label{3.2}
In this stage, we carefully design different injection modules based on the characteristics of each control signal and their relationship to achieve unified optical flow generation.
In detail, we categorize the injection modules into explicit and implicit injection based on the attributes of visual control signals. 
The explicit injection module is proposed to control signals that can be directly converted into sparse optical flow for one or some frames, such as arrow-based motion annotation on specific pixels.
The implicit injection is to incorporate control signals that are difficult to directly convert to pixel-level optical flow like camera trajectory.
Finally, since information in the reference image, such as semantic categories, is directly related to various control signals, the image is involved in both implicit and explicit injection methods. 
In the following, we will explain the detailed operations for explicit and implicit injection, and further discuss the unified control signal at this stage. 

\noindent\textbf{Explicit Injection.} 
As shown in \figref{fig:pipeline}, we explicitly convert different explicit control signals into initial sparse optical flow, and then apply a classical latent diffusion model~\cite{Rombach_Blattmann_Lorenz_Esser_Ommer_2022}, namely the Flow Generation Model (FGM), to transform it into dense optical flows.
For these signals like arrow-based motion annotation, we use the following pipeline for conversion.
Given a reference image, the user can label various motion trajectories on the image to represent the desired movements of objects and the environment. 
Take one trajectory for example, the trajectory can be regarded as a 2D point set, $\mathcal{M} \in \mathbb{R}^{P \times 2}=\left[\left(x_{0}, y_{0}\right),\left(x_{1}, y_{1}\right), \ldots,\left(x_{P-1}, y_{P-1}\right)\right]$, where $P$ is the point number. The sparse control points can be extracted from $\mathcal{M}$ with bicubic interpolation and then used to generate a point-wise sparse motion flow $F^{s}$ in the following equation, the same as MOFA-Video~\cite{niu2024mofa}, allowing us to guide the object motions and environmental changes effectively. 
\begin{equation}
\begin{array}{l}
F^{s}_{l-1}(x_{i},y_{i}) =\hat{\mathcal{T}}_{l}(x_{i},y_{i})-\hat{\mathcal{T}}_{0}(x_{i},y_{i})
\end{array}
\end{equation}
where $l\in\{1,2, \ldots, L-1\}$, $i$ denotes each pixel in the image. We also use CMP~\cite{Zhan_Pan_Liu_Lin_Loy_2019} to enhance sparse optical flows.
Theoretically, any visual control signals convertible to sparse optical flow through extraction~\cite{xu2023unifying} or generation~\cite{chatterjee2020sound2sight, yi2023focusflow}, such as audio~\citep{Ge2024OpFlowTalkerRA,chatterjee2020sound2sight}, videos, and object landmarks~\cite{fnevr},~\etc, can be input to FGM. 


\noindent\textbf{Implicit Injection.} 
For implicit control signals like camera trajectory condition, we adopt the progressive condition injection design of AnimateAnyone~\cite{hu2023animateanyone} and implicitly embed it into the FGM denoising process through the Camera Reference Model (CRM) progressively. The CRM employs a pre-trained image generation network based on the U-Net~\cite{ronneberger2015u} architecture (SD1.5\footnote{https://huggingface.co/stable-diffusion-v1-5/stable-diffusion-v1-5}) and is initialized with original weights. It integrates camera trajectories with the reference image to get multi-scale reference features at specific time steps through camera motion attention, which uses image latens as query, camera features as both the key and value in the original cross-attention part. Then these features are utilized to guide the generation of dense optical flow via the reference attention layer in the FGM the same as AnimateAnyone. 
In order to better describe the camera pose, we use Pl\"ucker embeddings~\citep{pluking_embedd} as the representation of camera trajectory. Given the extrinsic and intrinsic camera parameters $\mathbf{R}, \mathbf{t}, \boldsymbol{K}_{f}$ for the $f$-th frame, we derive a Pl\"ucker embedding $\ddot{\boldsymbol{p}}_{f, h, w}v \in \mathbb{R}^{6}$ for each pixel located at $(h, w)$. This embedding represents the vector from the camera center to the pixel's position as:
\begin{equation}
   \ddot{\boldsymbol{p}}_{f, h, w}=\left(\boldsymbol{t}_{f} \times \hat{\boldsymbol{d}}_{f, h, w}, \hat{\boldsymbol{d}}_{f, h, w}\right)
\end{equation}
\begin{equation}
\resizebox{0.9\linewidth}{!}{$
       \quad \hat{\boldsymbol{d}}_{f, h, w}=\frac{\boldsymbol{d}}{\left\|\boldsymbol{d}_{f, h, w}\right\|}, \quad \boldsymbol{d}_{f, h, w}=\boldsymbol{R}_{f} \boldsymbol{K}_{f}[w, h, 1]^{\top}+\boldsymbol{t}_{f}
$}
\end{equation}
Computing Pl\"ucker embedding for each pixel results in a representation $\ddot{\boldsymbol{P}} \in \mathbb{R}^{6 \times F \times H \times W}$ for a specified trajectory. To inject the trajectory representation into the reference motion network, we designed a trajectory encoder structurally similar to the camera encoder in CameraCtrl \citep{he2024cameractrl}. However, we improved the architecture: after each 2D ResNet block, we replaced the temporal attention with self-attention and output multi-scale trajectory features.



Through the combined use of explicit and implicit injection, we can effectively mitigate the incoherent pixel motion caused by different control signals.
In addition, we used Unimatch~\cite{xu2023unifying} to extract a high-quality optical flow from training videos as ground truth during training. Similar to Motion-I2V \cite{Shi_Huang_Wang_2024}, we trained a Flow Variational Autoencoder (VAE) to compress the optical flow of the pixel space into a latent flow space reducing computational resources. 


\subsection{Stage 2: Video Generation}
\label{3.3}
In the second stage, we aim to use the unified dense optical flow representation from the previous stage to guide the video generation model in creating a final video that aligns with the semantics of the reference image and annotations, as shown in \figref{fig:pipeline}. For the video generation model, we inherit from CogVideoX framework~\cite{yang2024cogvideox,hong2022cogvideo}. However, we introduce optical flow as a conditional guidance. Specifically, we use a flow encoder to encode the flows as the flow latent $z_{f}$. The flow encoder adopts four symmetrically arranged stages, respectively performing 2× downsampling and upsampling by the interleaving of ResNet block stacked stages achieving a 4× compression in the temporal dimension and an 8×8 compression in the spatial dimension similar to the 3D VAE encoder~\cite{yang2024cogvideox}.
Then we use a single basic Vision Transformer (ViT) block\cite{dosovitskiy2020image} to query video latents $z_{v}$ from the flow latents$z_{f}$ before calculating self-attention in the ViT block. 
\begin{equation}
    z^{\prime}_{v}=\operatorname{Attention}(\mathrm{Q}, \mathrm{~K}, \mathrm{~V})=\operatorname{Softmax}\left(\mathrm{QK}^{\mathrm{T}}\right) \mathrm{V},
\end{equation}
where $Q=W^{Q}z_{v}$,$K=W^{K}z_{f}$,$V=W^{V}z_{f}$. Notably, the flow feature maps only serve as key and value features. 
At the same time, the text prompts go through the text encoder using the google-research T5 model~\cite{roberts2022t5x}. The result of the transformer block is then concatenated with text embedding before going through the full-attention transformer block.
As shown in \figref{fig:pipeline}, we only train the optical flow encoder, input transformer block, and our video smoothing module (detailed in \secref{3.4}), keeping the parameters of other parts fixed to reduce the training difficulty. 

\subsection{Frequency Stabilization}
\label{3.4}
\begin{figure}[t]
    \centering
    \includegraphics[width=0.9\linewidth]{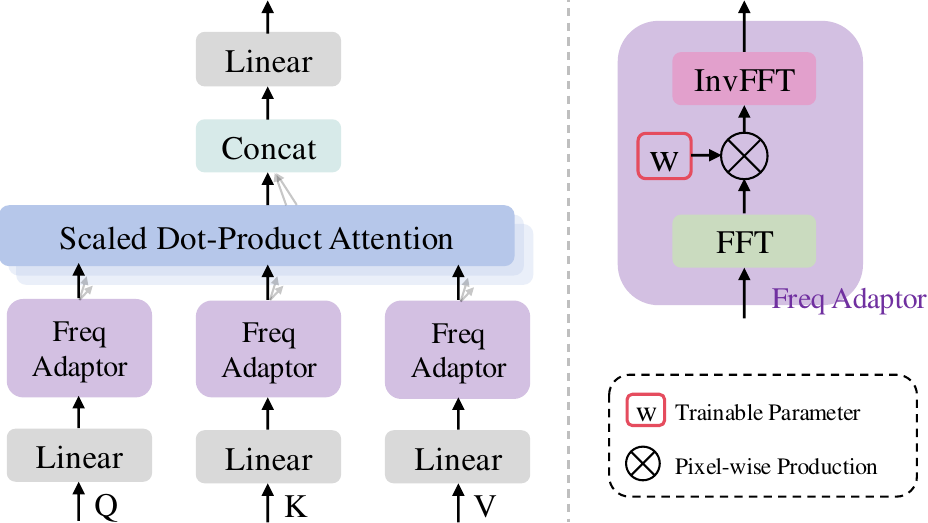}
    \caption{Video stabilization Module}
    \label{fig:video_stabilization}
\end{figure}

In the previous two subsections, we effectively incorporated a large amount of motion control into our network through a two-stage design, supporting the generation of significant motion variations.
In such cases, the corresponding optical flow may change drastically, making it prone to flickering and instability in the final video.
To solve this problem, we review the video generation task from the perspective of information encoding. 
Treating the generated video as a sequence of images, flickering typically occurs due to misalignment of features between frames.
This stems from the training of video generation models, where noise added to different frames at the same time step is independent. Despite temporal interactions among features, it remains challenging to prevent noise from negatively affecting the continuity and stability of video features.
This instability will greatly affect the video generation quality. 
Compared to temporal features, frequency-domain features can more directly reveal some essential video-level information from a different perspective with individual frequency components, which is important for suppressing flickering issues.
Thus, we adaptively modify the frequency-domain features extracted with the Fast Fourier Transform~(FFT)~\cite{cooley1967historical} to maintain temporal stability. 
Specifically, as shown in \figref{fig:video_stabilization}, we modify the attention mechanism in the Diffusion Transformer~(DiT)~\cite{Peebles_Xie_2022} architecture by first applying an FFT to each weight matrix to obtain its spectral features. We then multiply these features by a parameterized weight matrix $W$, followed by an inverse FFT~(InvFFT) to restore the original temporal-domain feature. This is then used to compute dot-product attention, ensuring the consistency of scene features along the temporal direction during video generation.

\subsection{Training Strategy}
\label{3.5}

We conducted experiments on a server equipped with $8\times$ NVIDIA Tesla A800 80G GPUs.
In the first stage, for optical flow generation, we primarily use the Real10K~\cite{Zhou_Tucker_Flynn_Fyffe_Snavely_2018} and DL3DV10K \cite{ling2024dl3dv} datasets for training. In the second stage, for video generation, we utilize the WebVid10M~\citep{Bain_Nagrani_Varol_Zisserman_2021} and OpenVid~\cite{nan2024openvid} datasets.
Both datasets are large and diverse, covering various aspects of daily life from multiple sources, ensuring strong generalization.

Currently, achieving large-scale camera trajectory control for video generation remains a significant challenge. 
One of the major difficulties is the limited availability of video data with camera trajectory poses. The datasets available are Real10K~\cite{Zhou_Tucker_Flynn_Fyffe_Snavely_2018} and DL3DV10K~\cite{ling2024dl3dv}, but both are primarily indoor or static scene datasets. The video model trained on these datasets is unsuitable for dynamic scenes, while the generation is prone to failure. 
Another difficulty is that the dynamic video datasets available rarely contain pose information due to the difficulty of camera pose estimation in dynamic scenes using structure-from-motion (SfM) methods like COLMAP~\cite{Schonberger_Frahm_2016}. 
So, we organize and augment the data, and we further decompose the training of the first stage to achieve multi-condition controllable network training with limited data.
Through careful search, we found that many videos on OpenVid~\cite{nan2024openvid} are shot from fixed camera positions, which can serve as a good starting point to boost dynamic training. We select a batch of videos with roughly fixed camera positions by evaluating the motion magnitude of the global optical flow, totaling around 10,000 videos. Our model is to first train the initial model on the Real10K dataset, then set the camera viewpoint of this batch of videos to be fixed at the origin, and subsequently train our model using the selected dynamic videos.

Given that the optical flow data required for the second stage is directly sourced from the video, both stages can be trained independently, with connection only needed during the inference process. Additionally, we apply noise to the optical flow in the training of the second stage to enhance the learning capability of the video generation model.

\section{Experiments}
\begin{figure*}[ht]
    \centering
    \includegraphics[width=\textwidth]{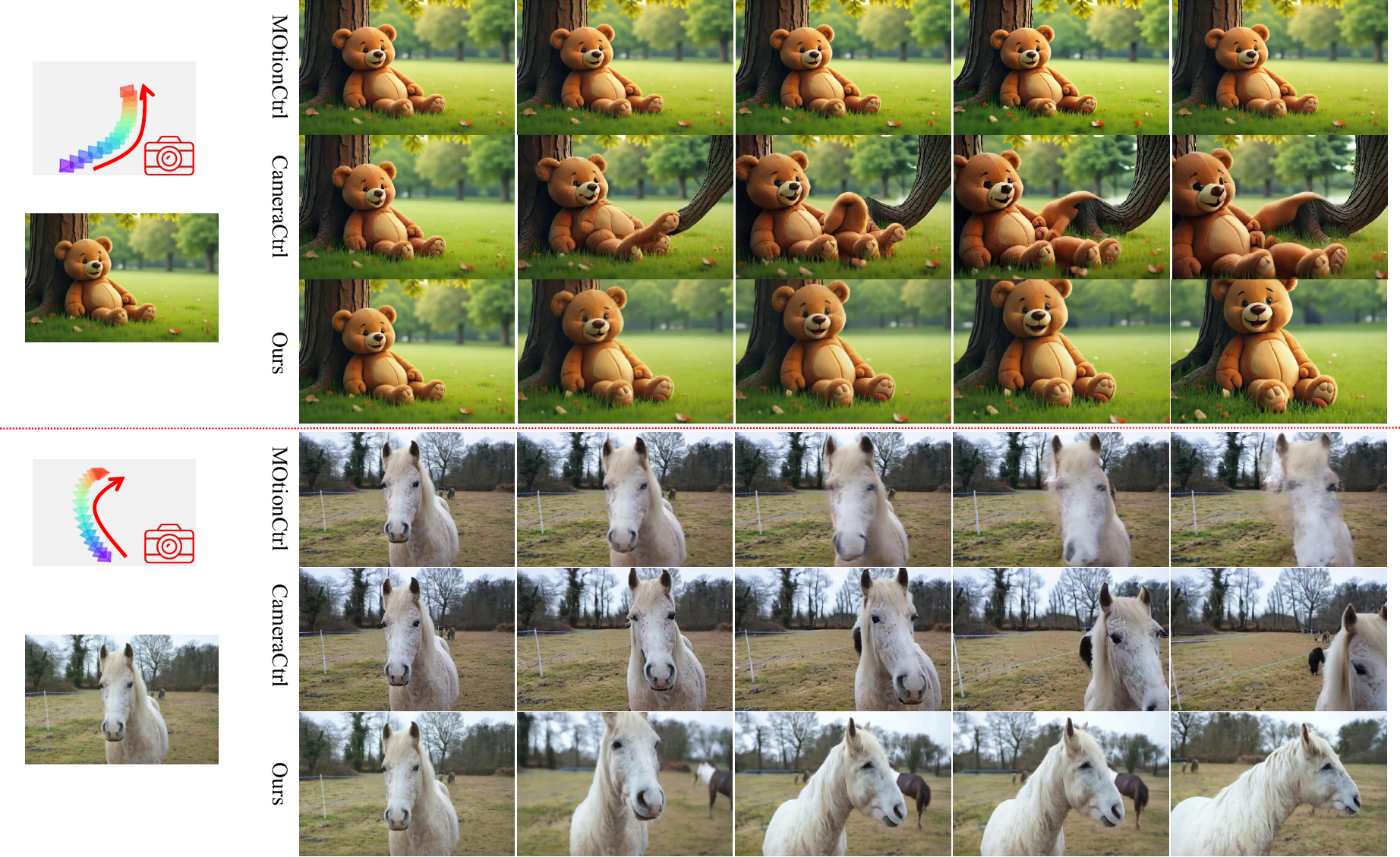}
   \vspace{-1.7em}
    \caption{Camera trajectory comparison with other trajectory-based methods}
    \label{camera_pose}
\end{figure*}

\begin{figure*}[ht]
    \includegraphics[width=\textwidth]{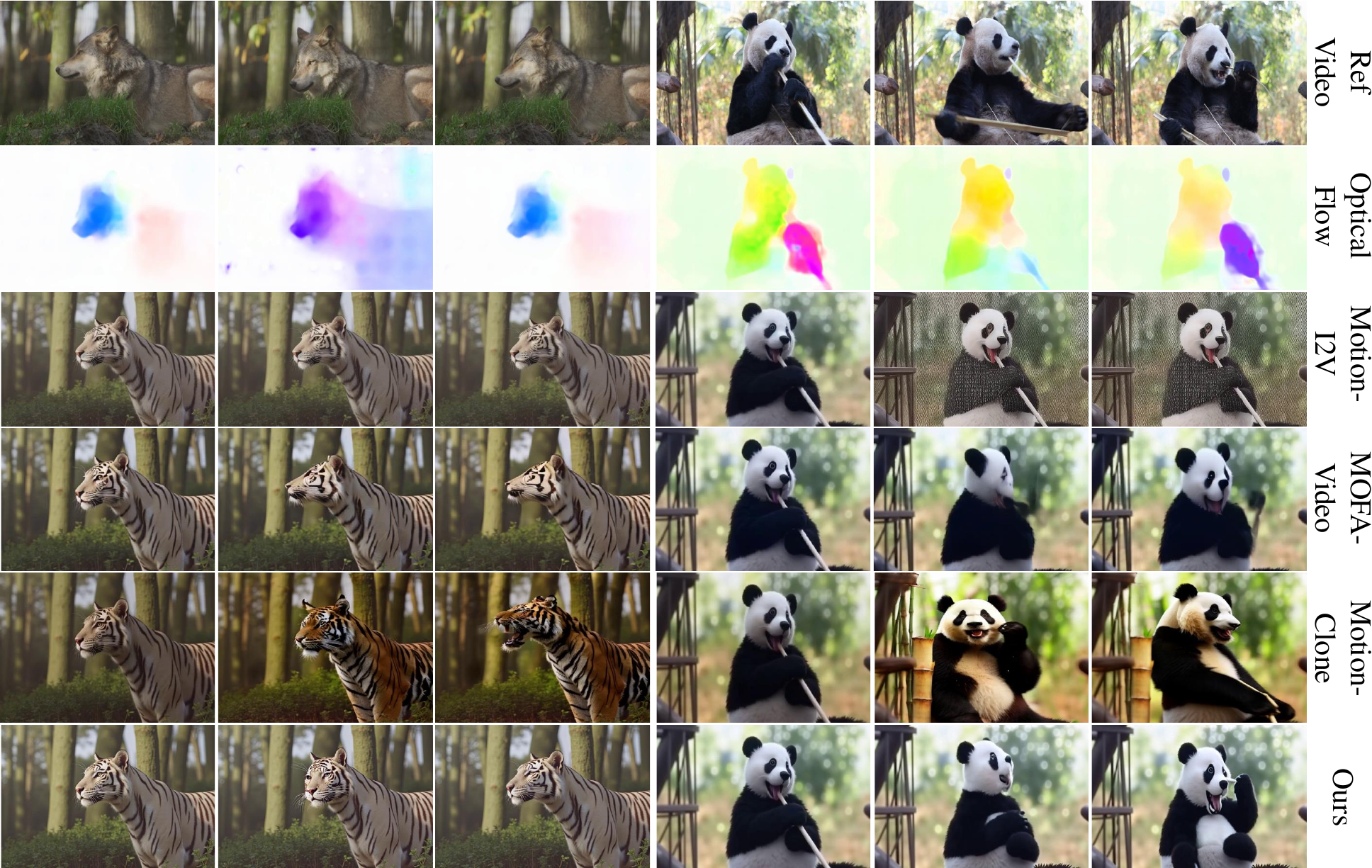}
   \vspace{-1.7em}
    \caption{Motion Transfer comparison with state-of-the-art methods.
   }
   \label{motionclone}
\end{figure*}

\begin{figure}[ht]
    \centering
    \includegraphics[width=\linewidth]{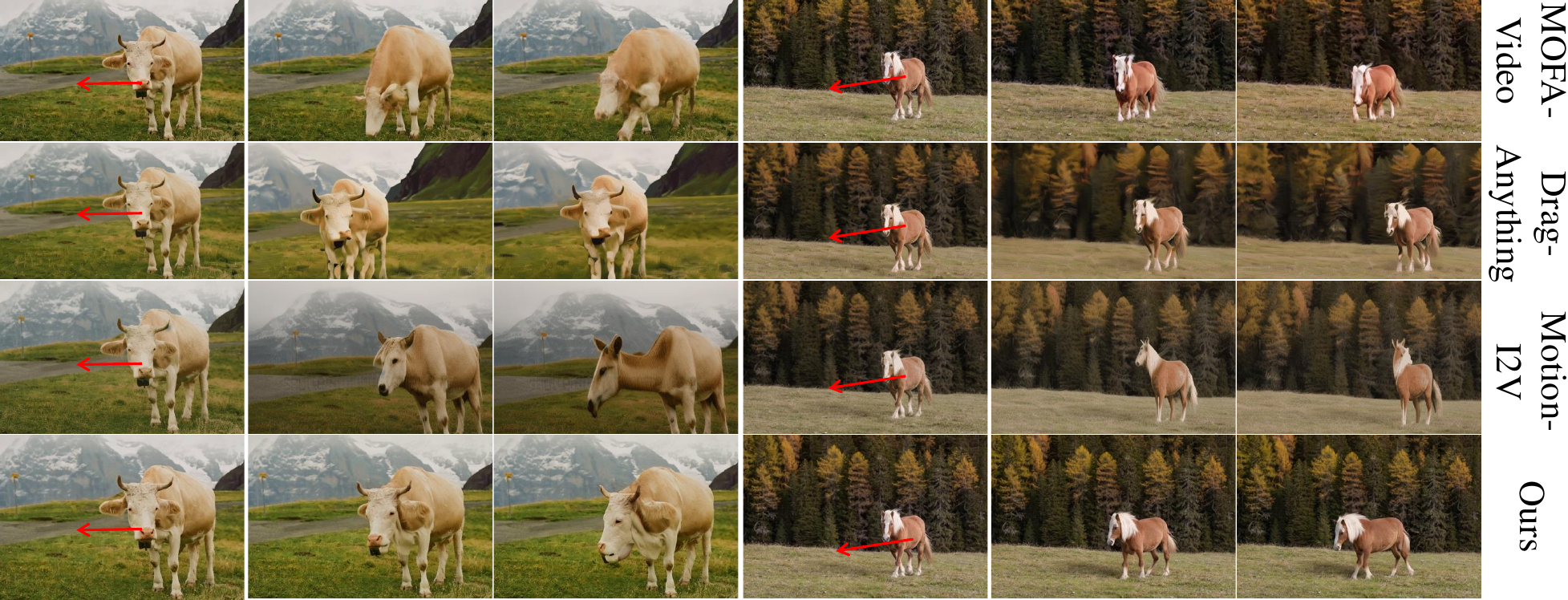}
   \vspace{-1.7em}
    \caption{Users drag animation comparison with other animation methods.}
    \label{fig:compare_animate}
\end{figure}

We evaluate our method through quantitative metrics that confirm both the generation quality and its alignment with control signals, alongside visualizing the generated results for further qualitative comparison.

\subsection{Image-to-Video Generation Ability.}
\begin{figure}[ht]
    \includegraphics[width=\linewidth]{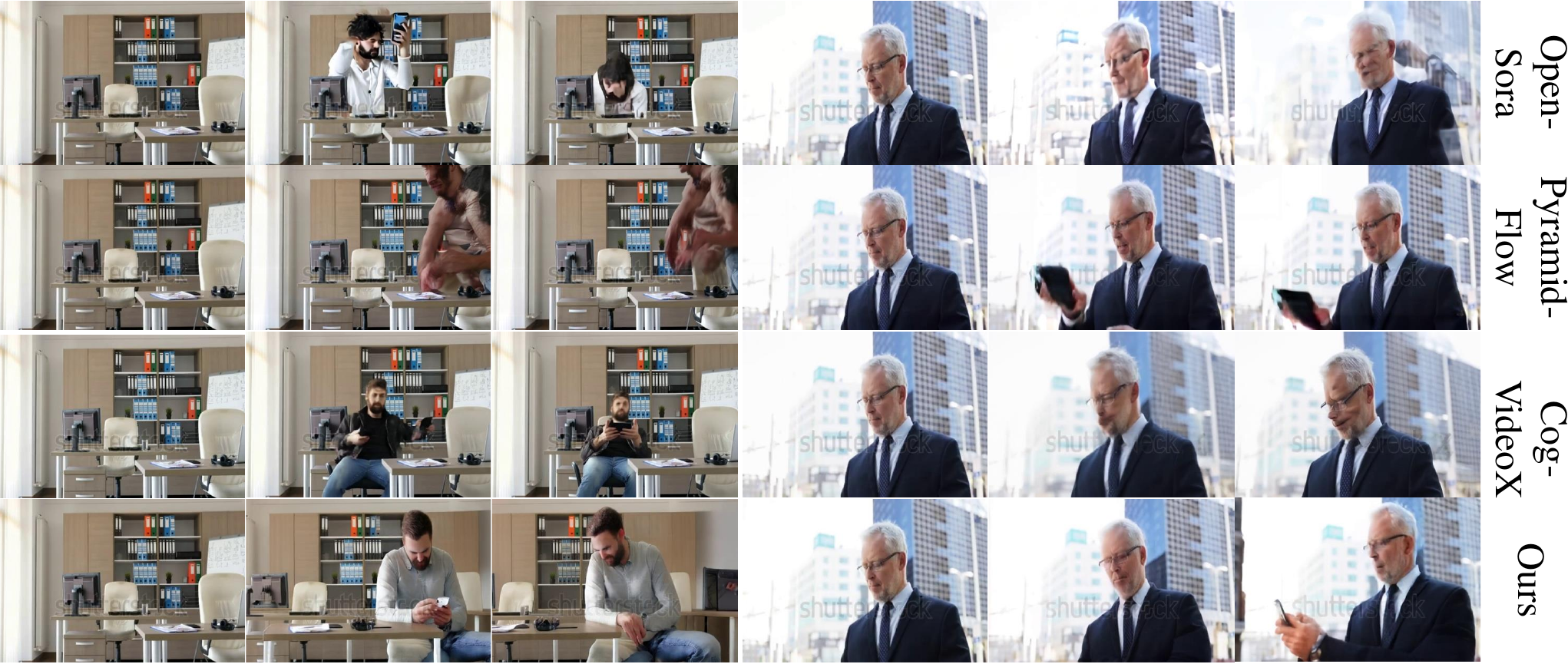}
   \vspace{-1.7em}
    \caption{Image to video generation comparison with current state-of-the-art methods.
   }
   \label{video_quality_image}
   \vspace{-1em}
\end{figure}

As shown in \tabref{video_quality}, four classical image-level quality metrics, including Fr\'{e}chet Inception Distance (FID)~\citep{Heusel_Ramsauer_Unterthiner_Nessler_Hochreiter_2017},  SSIM~\citep{Wang_Bovik_Sheikh_Simoncelli_2004}, PSNR~\citep{Hore_Ziou_2010} and LPIPS~\citep{Zhang_Isola_Efros_Shechtman_Wang_2018}, are used to evaluate the quality of the generated video frames with Motion-I2V~\cite{Shi_Huang_Wang_2024}, MOFA-Video~\cite{niu2024mofa}, DynamiCrafter~\cite{xing2023dynamicrafter}, CogVideoX\cite{yang2024cogvideox}, PyramidFlow~\cite{jin2024pyramidal},
and OpenSora~\cite{opensora} , and video-level metric Fr\'{e}chet Video Distance (FVD)~\citep{Unterthiner_Steenkiste_Kurach_Marinier_Michalski_Gelly_2018} is applied to assess video-level quality, similar to previous video generation methods~\citep{ Wu_Ge_Wang_Lei_Gu_Hsu_Shan_Qie_Shou_2022, Khachatryan_Movsisyan_Tadevosyan_Henschel_Wang_Navasardyan_Shi, Cai_Chan_Peng_Shahbazi_Obukhov_Gool_Wetzstein_2022, Ceylan_Huang_Mitra, Cai_Ceylan_Gadelha_Huang_Wang_Wetzstein_2023}. 
Following CogVideo~\cite{yang2024cogvideox} and PyramidFlow~\cite{jin2024pyramidal}, we employed several metrics from VBench~\cite{huang2023vbench} to evaluate the Subject Consistency~(SubC), Motion Smoothness~(MoS) and Aesthetic Quality~(AesQ) of Our Video as shown in \tabref{video_smooth} and \figref{video_quality_image} on OpenVid~\citep{nan2024openvid} and WebVid~\citep{Bain21} datasets.
With optical flow guidance, our methods can achieve better performance especially when the generated video contains human motions and animal motions.

\subsection{Control Signals Driven I2V Generation.}

\noindent\textbf{Camera Trajectory.}
The distance between predicted and ground-truth camera trajectories is used to measure the camera alignment. Here, we use ParticleSfM\cite{Zhao_Liu_Guo_Wang_Liu_2022}, VggSfM\cite{wang2024vggsfm} and DUSt3R\cite{DUSt3R_cvpr24} to evaluate our camera trajectory with CameraCtrl\cite{he2024cameractrl} and MotionCtrl\cite{wang2024motionctrl} on basic trajectory (sample every 8 frames) and difficult trajectory (sample every max frame we can sample) in Real10K~\cite{re10k} shown in \tabref{camera_pose} and \figref{camera_pose}.
Specifically, we estimate the camera trajectories for both the generated and real videos using the same methods to eliminate the potential scale differences caused by different Structure-from-Motion~(SfM) techniques.
And, we evaluate the quality of these trajectories by evaluating the scale and differences in the rotation and translation parameters of the camera matrix using \emph{rotation error} and \emph{translation error}, as outlined in~\citet{wang2024motionctrl, he2024cameractrl}.

\begin{table}[t]
\caption{Quantitative comparisons~(Pose got by DUSt3R, VggSfM, and ParticleSfM). We compare against prior works on basic trajectory and random trajectory respectively. T-Err, R-Err represent \emph{translation error} and \emph{rotation error}.}
\vspace{-8pt}
\label{camera_pose}
\resizebox{1.0\linewidth}{!}{
\begin{tabular}{cc@{\hspace{0.1cm}}c@{\hspace{0.1cm}}c@{\hspace{0.1cm}}c@{\hspace{0.1cm}}c@{\hspace{0.1cm}}c@{\hspace{0.1cm}}c@{\hspace{0.1cm}}c@{\hspace{0.1cm}}c@{\hspace{0.1cm}}c@{\hspace{0.1cm}}c@{\hspace{0.1cm}}c@{\hspace{0.1cm}}}
\toprule
\multicolumn{1}{l}{} & \multicolumn{6}{c}{Basic Trajectory}                                                  & \multicolumn{6}{c}{Difficult Trajectory} \\ \cmidrule(r){2-7} \cmidrule(l{1pt}){8-13}
\multicolumn{1}{l}{} & \multicolumn{2}{c}{DUSt3R}      & \multicolumn{2}{c}{VggSfM}  & \multicolumn{2}{c}{ParticleSfM} & \multicolumn{2}{c}{DUSt3R}      & \multicolumn{2}{c}{VggSfM}  & \multicolumn{2}{c}{ParticleSfM} \\ 
\cmidrule(r{1pt}){2-3} \cmidrule(l{1pt}r{1pt}){4-5} \cmidrule(l{1pt}r){6-7} \cmidrule(l{1pt}r{1pt}){8-9} \cmidrule(l{1pt}r{1pt}){10-11} \cmidrule(l{1pt}){12-13}
                     & T-Err$\downarrow$ & R-Err$\downarrow$ & T-Err$\downarrow$ & R-Err$\downarrow$ & T-Err$\downarrow$ & R-Err$\downarrow$ & T-Err$\downarrow$ & R-Err$\downarrow$ & T-Err$\downarrow$ & R-Err$\downarrow$ & T-Err$\downarrow$ & R-Err$\downarrow$  \\ \midrule
CameraCtrl            &0.090 &0.300 &1.405 &0.177 &2.277 &0.825   &0.082 &0.306  &1.559 &0.144 & 2.172 &0.722                            \\
MotionCtrl       &0.057 &0.233 &1.324 &0.258 &1.811 &0.868   &0.060 &0.267 &0.875 &0.137    &2.424 &0.756                                  \\ 
Ours          &\first{0.041} & \first{0.159}  & \first{1.036} & \first{0.125} & \first{1.648} & \first{0.685} & \first{0.053} & \first{0.203} & \first{0.447} & \first{0.119} & \first{2.042} & \first{0.572}                 \\ \bottomrule

\end{tabular}
}
\end{table}

\begin{figure}[ht]
    \centering
    \includegraphics[width=\linewidth]{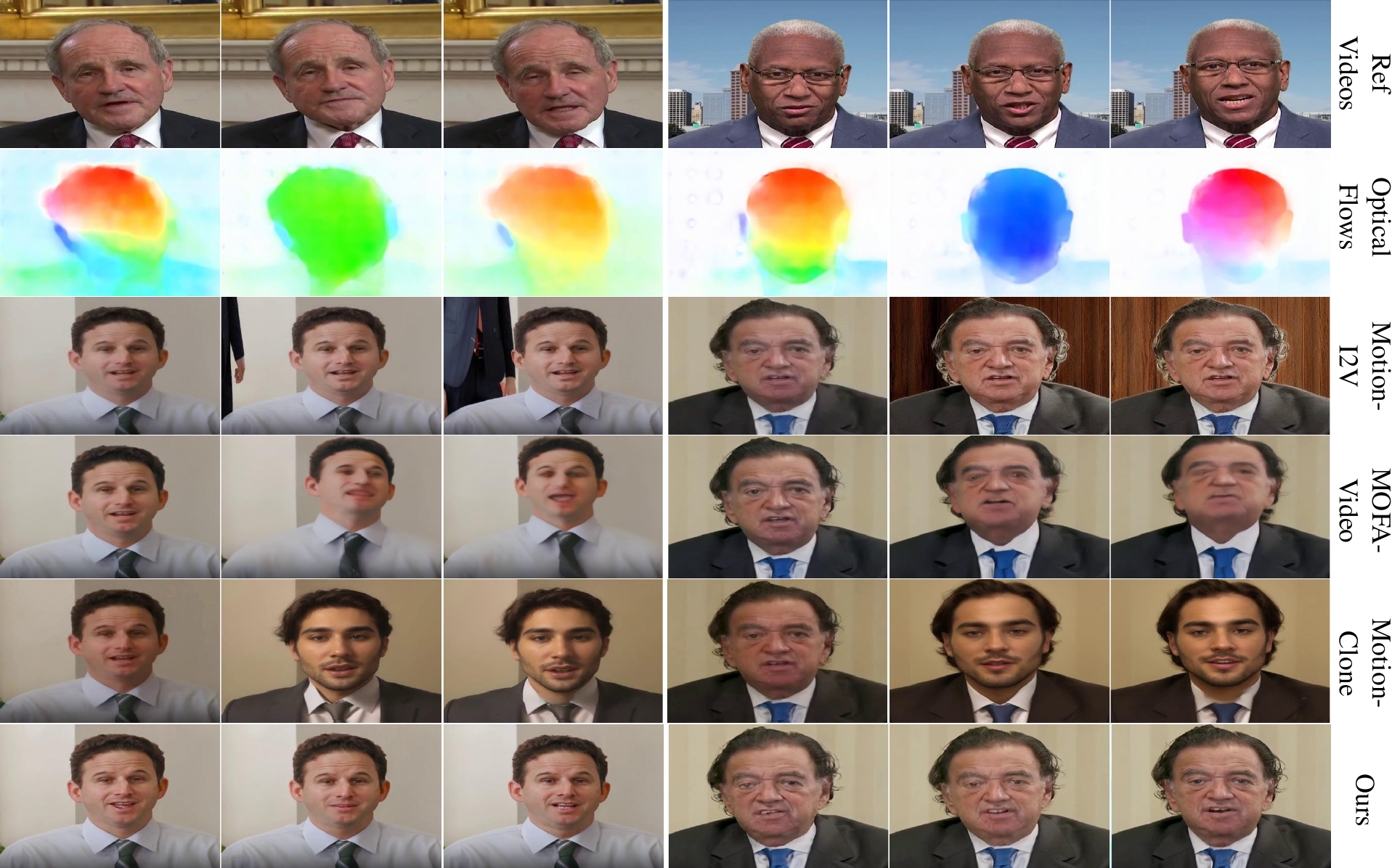}
   \vspace{-2em}
    \caption{Human face animation with optical flow extracted from reference video}
    \label{human_face}
\end{figure}

\begin{table*}[th]
\small
\centering
\caption{Video quality comparison.}
\label{video_quality}
\resizebox{0.8\linewidth}{!}{
\begin{tabular}{ccccccccccc}
\toprule
               & \multicolumn{5}{c}{webvid}    & \multicolumn{5}{c}{OpenVid}       \\ \cmidrule(rl){2-6} \cmidrule(l){7-11}
               & LPIPS$\downarrow$ & PSNR$\uparrow$ & SSIM$\uparrow$ & FID$\downarrow$ & FVD$\downarrow$ & LPIPS$\downarrow$ & PSNR$\uparrow$ & SSIM$\uparrow$ & FID$\downarrow$ & FVD$\downarrow$ \\ \midrule
Motion-I2V     &0.375  &16.14  &0.487 &94.77 &720  &0.329   &15.28   &0.488 &72.14 &704 \\
MOFA-Video     &0.351  &18.43   &0.603 & \cellthird 57.12 &524  &0.300   &19.64   &0.655 &52.66 &654 \\
DynamiCrafter  &0.268  &18.56   &0.532 &63.73 &685  &0.393   &13.83   &0.402 &59.61 &751 \\
CogVideoX+image  & \cellsecond 0.147  & \cellthird 24.22 & \cellthird 0.762 &59.20 & \cellthird 486  &0.164  & 22.61  & \cellthird 0.762 & \cellthird 43.29 &547 \\
Pyramid-Flow   & \cellthird 0.152  & \cellsecond 24.99   & \cellsecond 0.792 
 & \cellsecond 55.78  & \cellsecond 470  & \cellthird 0.122   & \cellsecond 23.37   & \cellsecond 0.789 & \cellsecond 39.48 & \cellsecond 453 \\
Open-Sora     &0.179  &23.21   &0.725 &58.33 &552  & \cellsecond 0.117   & \cellthird 22.78   &0.760 &44.48 & \cellthird 512 \\
Ours            & \cellfirst 0.135  & \cellfirst 25.22  & \cellfirst 0.810 & \cellfirst 48.11  &  \cellfirst 380   & \cellfirst 0.113 & \cellfirst 25.04  &  \cellfirst 0.836 & \cellfirst 33.12 & \cellfirst 322        \\ \bottomrule
\end{tabular}
}
\end{table*}

\begin{table}[th]
\caption{Video consistency quality comparison. SubC: Subject Consistency; MoS: Motion Smoothness; AesQ: Aesthetic Quality.}
\label{video_smooth}
\resizebox{1.0\linewidth}{!}{
\begin{tabular}{ccccccc}
\toprule
               & \multicolumn{3}{c}{webvid}    & \multicolumn{3}{c}{OpenVid}       \\ \cmidrule(rl){2-4} \cmidrule(l){5-7}
               & SubC $\uparrow$ & MoS $\uparrow$ & Aesq $\uparrow$ & SubC $\uparrow$ & MoS $\uparrow$ & Aesq $\uparrow$  \\ \midrule
DynamiCrafter   &0.832  &0.958 &0.443    &0.910 &0.964 &0.536\\
CogVideoX+image  &0.855 &0.984  &0.443    &0.929 &0.987&0.567      \\
Pyramid-Flow     &0.906  &\first{0.991} &0.438    &0.941 &0.991 &0.537    \\
Open-Sora       &0.897 &0.989 &0.438    &0.954 &0.990 &0.524 \\
Ours            & \first{0.928}  & \first{0.991} & \first{0.474}      & \first{0.971} & \first{0.993} & \first{0.600}       \\ \bottomrule
\end{tabular}
}
\end{table}

\noindent\textbf{User Arrow Annotation.}
As shown in \figref{fig:intro}, we can turn any kind of user drags into corresponding optical flows, which are then treated as the unified guidance for the final video generation. 
For this part, we compare with current state-of-the-art user drag animation methods MOFA-Video~\cite{niu2024mofa}, DragAnything~\cite{wu2024draganything}, Motion-I2V~\cite{Shi_Huang_Wang_2024} shown in \figref{fig:compare_animate}. Our method can achieve more stable and consistent video generation on specific user drags. More results can be seen on the anonymous project webpage. 

\noindent\textbf{Reference Video.}
\label{ExpGeneralization}
We demonstrate the capability of our Stage 2 in generating animations driven by reference videos, where a dense optical flow can be extracted. First, we test the case that the optical flow and given images are well aligned. Specifically, we generated the reference image by replacing or stylizing the subject in the first frame of the video as the reference image~\cite {rout2024rfinversion}. As shown in \figref{motionclone}, it can be seen that Motion-I2V lacks sensitivity to a wide range of motions. Although MotionClone and MOFA-Video can achieve significant video motions, they result in style inconsistency and artifacts. Our generated results maintain significant motion alignment without a skeleton or facial keypoint extraction, while achieving optimal subject consistency.
To better evaluate the generalization, we further experiment on the facial replacement task with the setting that the image and the unified optical flow are not perfectly aligned. The dense optical flow here inputted to Stage 2 is directly extracted from another facial motion video. As shown in \figref{human_face}, our Stage 2 can tolerate some misalignments while still performing effectively, producing consistent expressions and lip motions. This provides our method with greater flexibility and robustness.

\subsection{Ablation Study And Analysis}

\begin{table}[]
\caption{Ablation study.}
\label{ablation}
\vspace{-10pt}
\label{Tab.AblationStudy}
\begin{center}
\resizebox{1.0\linewidth}{!}{
\begin{tabular}{cccccccc}
\toprule
                                             & \multicolumn{5}{c}{Visual Quality} & \multicolumn{2}{c}{Trajectory Alignment} \\  \cmidrule(lr){2-6} \cmidrule(l){7-8}
                                             & LPIPS$\downarrow$   & PSNR$\uparrow$   & SSIM$\uparrow$  & FID$\downarrow$  & FVD$\downarrow$  & TransErr$\downarrow$            & RotErr$\downarrow$           \\ \midrule
\multicolumn{1}{c}{Camera embedding}&0.401  &14.22   &0.531 &52.46 &346  &0.551   &0.048    \\
\multicolumn{1}{c}{ControlNet-Like}    &0.400  &14.21   &0.528 &50.96 &356  &0.737   &0.050   \\
\midrule
\multicolumn{1}{c}{w/o FS}    &0.241  &17.88   &0.615 &46.85 &311  &0.671   & 0.059    \\
\multicolumn{1}{c}{w/o noise}    &0.228  &19.32   &0.654 &49.38 &474  &0.425   &0.048   \\
\multicolumn{1}{c}{Full Model}   & \first{0.142}  & \first{23.22}   & \first{0.796} & \first{41.67}  & \first{168}  & \first{0.354}   & \first{0.047}               \\ \bottomrule
\end{tabular}
}
\vspace{-2em}
\end{center}
\end{table}

\noindent\textbf{Setups.} To verify the effectiveness of the components in our video generation pipeline, we designed several sets of ablation experiments on Real10K~\cite{re10k}. 
(1) The multi-frame camera encoding is added to the latent variables and used as input to the DiT blocks. (2)We replicate the first half of the blocks of FGM as a reference network and then add the output of each block of the reference network to the output of the corresponding original block, like ControlNet. (3) and (4) both use the globally estimated optical flow from the video data as input, with the distinction that (3) removes Frequency Stabilization (FS), while (4) does not apply noise before feeding the global optical flow into the Flow Encoder.

\noindent\textbf{Analysis.} 
As shown in \tabref{Tab.AblationStudy}, 
From the first two rows, we can see the superiority of using a unified optical flow representation, which surpasses other camera control signal guidance methods in terms of visual quality and camera rule prediction. The third row reflects that not using Frequency Stability during the training process leads to a significant drop in performance. However, it is still better than the other two camera signal guidance methods. The fourth row illustrates that the noise application conducted before feeding the Optical Flow into the Flow encoder effectively enhances the robustness of the generation.


\section{Conclusion}

In this paper, we present a unified controllable video generation approach enabling precise and consistent video manipulation across various conditions.
We unified Flow as a joint control signal by converting diverse visual control signals (e.g., object motion, camera motion) into a joint optical flow representation. Then the unified flows are used to guide the final video generation. This strategy reduces the complexity of handling multiple, isolated control signals and promotes consistency in the generated video. In addition, we propose a frequency-based stabilization module to preserve the key features in the frequency domain and reduce the flickering issues caused by large-scale motion. Experiments demonstrate that this two-stage pipeline can control the video generation precisely and have impressive generalization capabilities.

{
    \small
    \bibliographystyle{ieeenat_fullname}
    \bibliography{main}
}


\end{document}